\documentclass[conference]{IEEEtran}
\IEEEoverridecommandlockouts

\usepackage{cite}
\usepackage{amsmath,amssymb,amsfonts}

\usepackage{graphicx}
\graphicspath{{figures/}}
\usepackage{float} 
\usepackage{textcomp}
\usepackage{float}

\usepackage{array}
\usepackage{multirow}
\usepackage{arydshln}

\usepackage{graphicx,subcaption}
\usepackage{wrapfig}
\usepackage{xcolor}
\usepackage{tikz,pgfplots}
\usetikzlibrary{arrows}

\usepackage[acronym]{glossaries}

\usepackage{algorithm}
\usepackage{algpseudocode}
\usepackage{algorithmicx}

\usepackage{caption}
\usepackage[numbers]{natbib}
\usepackage[colorinlistoftodos]{todonotes}

\usepackage{hyperref} 
\hypersetup{
    linkbordercolor={0 0 1}
}

\usepackage{xcolor}
\def\BibTeX{{\rm B\kern-.05em{\sc i\kern-.025em b}\kern-.08em
    T\kern-.1667em\lower.7ex\hbox{E}\kern-.125emX}}

\begin{document}

\title{Collaborative Ground-Aerial Multi-Robot System for Disaster Response Missions with a Low-Cost Drone Add-On for Off-the-Shelf Drones \\ 

\vspace{-12pt}

}

\author{Shalutha Rajapakshe\IEEEauthorrefmark{1},
Dilanka Wickramasinghe\IEEEauthorrefmark{1},
Sahan Gurusinghe\IEEEauthorrefmark{1},
Deepana Ishtaweera\IEEEauthorrefmark{1},\\
Bhanuka Silva\IEEEauthorrefmark{1},
Peshala Jayasekara\IEEEauthorrefmark{1},
Nick Panitz\IEEEauthorrefmark{2},
Paul Flick\IEEEauthorrefmark{2}, and
Navinda Kottege\IEEEauthorrefmark{2}\\
        
        \IEEEauthorrefmark{1}Department of Electronic and Telecommunication Engineering, University of Moratuwa, Sri Lanka \\
        \IEEEauthorrefmark{2}Robotics and Autonomous Systems Group, CSIRO, Pullenvale, QLD 4069, Australia\\

\vspace{-27pt}

        }

\maketitle

\begin{abstract}

In disaster-stricken environments, it’s vital to assess the damage quickly, analyse the stability of the environment, and allocate resources to the most vulnerable areas where victims might be present. These missions are difficult and dangerous to be conducted directly by humans. Using the complementary capabilities of both the ground and aerial robots, we investigate a collaborative approach of aerial and ground robots to address this problem. With an increased field of view, faster speed, and compact size, the aerial robot explores the area and creates a 3D feature-based map graph of the environment while providing a live video stream to the ground control station.  Once the aerial robot finishes the exploration run, the ground control station processes the map and sends it to the ground robot. The ground robot, with its higher operation time, static stability, payload delivery and tele-conference capabilities, can then autonomously navigate to identified high-vulnerability locations. We have conducted experiments using a quadcopter and a hexapod robot in an indoor modelled environment with obstacles and uneven ground. Additionally, we have developed a low-cost drone add-on with value-added capabilities, such as victim detection, that can be attached to an off-the-shelf drone. The system was assessed for cost-effectiveness, energy efficiency, and scalability.


\end{abstract}

\vspace{-1pt}

\section{Introduction}

\vspace{-1pt}

Disaster situations typically require immediate responses to search and rescue victims, prevent post-disaster scenarios, and assess damages. However, hazardous conditions such as debris, toxic gases, radioactive materials, etc. can impede human rescuers. Consequently, there is a growing interest in developing robotics technologies to aid in these types of crisis circumstances. In disaster scenarios, multi-robot approaches have advantages over single-robot approaches \cite{2020litPaper}\cite{lit_paper1}\cite{Darpa_paper1}\cite{Darpa_paper2} as robots can perform task distribution, exchange information, and complete tasks more efficiently, resulting in faster completion times and reduced energy consumption.

Aerial robots are fast and offer a large field of view, but have limited flight time. Ground robots, on the other hand, are optimized for longer missions, can carry larger payloads, and actively interact with the environment, although their field of view can be obstructed. 
To optimize the use of both robot types, we propose a system\footnote{A project video demonstration can be found at: \url{https://www.youtube.com/watch?v=jzN4bU5uXSQ}} where initially, an aerial robot equipped with a custom add-on will explore and map an environment while identifying targets or human victims. During this process, the live visual feed can be observed. Then, the hexapod robot will utilise that map to autonomously reach the identified targets, inspect and interact with them, and deliver medi-packs. Additionally, The hexapod robot is equipped with tele-conferencing equipment to communicate with victims, which is an essential aspect in disaster scenarios.

\vspace{-1pt}

\section{Related Work} 

\vspace{-2pt}

D. Chatziparaschis et al. have presented a ground-aerial collaboration in \cite{lit_paper1} using a DJI Matrice 100 quadcopter and Aldebaran Nao humanoid robot. The approach involves the humanoid robot planning its own paths using a map provided by the unmanned aerial vehicle (UAV). The UAV creates a 3D map by collecting depth information through a multi-stereo camera system, followed by an octree mapping mechanism. In this approach, the humanoid can detect victims, but the UAV is unable to do so, due to possible occlusions from the multi-stereo camera system's placement structure. Moreover, the humanoid lacks mapping sensors or algorithms, and relies solely on the UAV vision system, which transfers its pose with respect to the UAV with the help of an AR marker mounted on top of the humanoid. Therefore, the approach does not effectively utilize the capabilities of both robots.

In \cite{lit_paper2}, an Extended Kalman Filter (EKF) based framework that combines data from a UAV (DJI Matrice 100 equipped with DJI Guidance module with a downward looking sensor unit) and an unmanned ground vehicle (UGV) has been introduced for target searching in complex environments. The UAV first  perform a surveying task to identify the target coordinates and then sends them to the UGV. The UGV is a custom-designed platform equipped with four mecanum wheels. UGV moves towards the target while avoiding obstacles after the UAV completes mapping and returns to the initial position. Then, UAV autonomously tracks the UGV using a marker to keep it within the UAV's camera view while following the path towards the goal. The requirement to mark targets with ArUco markers may not be suitable in cluttered environments. The UGV's localization heavily relies on a compass module, which may be affected in areas with magnetic interference.

An aerial robot which provides a bird's-eye view for the ground robot (KUKA youBot) has been introduced in \cite{lit_paper7} for search and rescue tasks. Leveraging the bird's eye view, the ground robot will plan its path to goal locations using A$^{*}$ algorithm, while removing obstacles using its 5-DOF manipulator with a two-finger gripper. The use of AprilTags to identify obstacles and the assumption of obstacles being stationary may limit the practical application of this approach in real world scenarios.

The system in \cite{lit_paper3} uses aerial and ground robots to map a multi-story environment collaboratively. A Pelican quadcopter equipped with an Inertial Measurement Unit (IMU) and two tracked robots, Kenaf and Quince were used in the approach. Initially, Kenaf is tele-operated through the multi-story environment while collecting sensory data. Then, the Quince which carries a tele-operated quadcopter and an automated helipad will be released to access locations inaccessible by Kenaf. After Kenaf reaches the locations, the helipad is remotely opened and the quadcopter autonomously takes off and continues the mapping process. Upon completion, the quadcopter is signalled to land autonomously on the helipad. This approach takes a longer time to map an area as the ground robot maps first.

\vspace{2pt}

The approach in \cite{lit_paper6} uses a UAV with a downward-facing monocular camera and a steering-based wheeled robot. The UAV measures the obstacles' poses relative to the UGV to build a global map that includes both UGV and obstacles with the help of an EKF. However, UGV must be within UAV's view, limiting the robots' independent navigation.

\vspace{2pt}

Multiple approaches use elevation maps for collaborative localization and navigation. In \cite{lit_paper4}, a drone equipped with a camera and IMU creates an elevation map, used as a reference for the quadrupedal ground robot, StarlETH. The localization method is independent of viewpoints and both robots can create elevation maps with onboard sensors. In \cite{lit_paper5}, a hexacopter collects visual data of an unknown area through a monocular camera, which is used to generate a probabilistic elevation map. Then the ground robot uses it to plan a traversable path by interpreting the elevation map as a traversability map. Moreover, the laser range sensor on the ground robot continuously updates the maps as it moves to plan new paths with updated data.

\vspace{2pt}

While the methodology proposed in \cite{2020Paper} for collaborative exploration is promising, it may not be suitable for disaster environments due to highly specific hardware requirements, high costs, and limitations of used ground robots to operate in cluttered environments. It is essential to consider the specific requirements of disaster response and recovery when assessing the suitability of robotic systems. 

\vspace{2pt}

Many of the used aerial robots in these type of collaborative methods are either off-the-shelf drones with specific hardware modifications or custom-built drones. This can make it difficult for users to switch to a different type of aerial robot if needed. To overcome this particular limitation, our method incorporates a cost-effective custom hardware add-on that is equipped with essential sensory devices. This enables a seamless attachment to any off-the-shelf drone that has the capacity to accommodate additional weight. This makes our approach highly scalable and adaptable to a wide range of different aerial robots. We chose a six-legged robot as our ground robot  based on an evaluation with other mobile ground robot types. The criteria include speed, traversing capabilities, mechanical complexity, disturbance to the ground, stability on uneven terrains, and adaptability to different terrain conditions.  

\vspace{-2pt}

\section{Methodology}

\vspace{-1pt}

This section is divided into four subsections. Section \ref{sysover} describes the high-level overview of the proposed system, including the used components and communication theories. Section \ref{droneaddon} discuss the details of the custom-designed drone add-on along with the mapping and victim detection aspects. Methodologies behind the six-legged robot and the tele-conferencing feature are described in detail in section \ref{hexapod}. Finally, section \ref{collab} provides details in regards to how the collaboration is achieved between the ground and aerial robots.

\vspace{-3pt}

\subsection{System Overview}\label{sysover}

\vspace{-10pt}

\begin{figure}[H]
  \centering
  \includegraphics[width=1\linewidth]{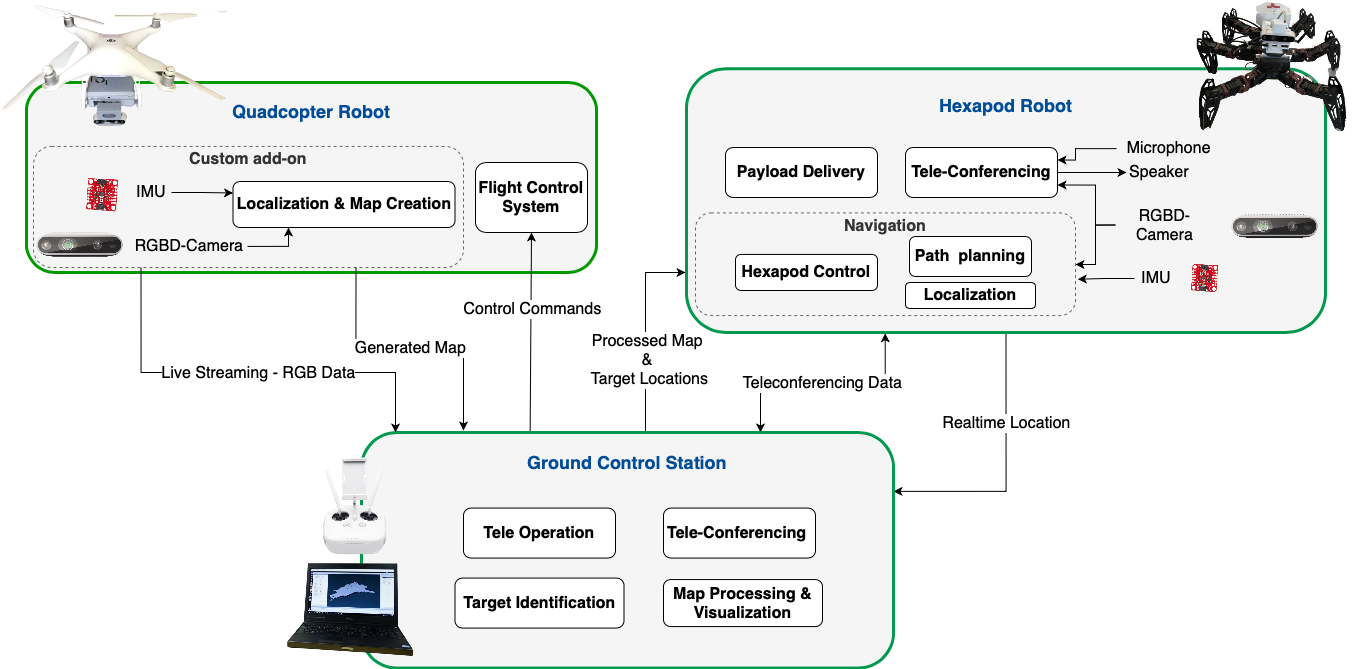} 
  \vspace{-5pt}
  \caption{System Architecture}
  \label{fig:system-architecture}
\end{figure} 

\vspace{-10pt}

The high-level architecture of the collaborative system is comprised of three main components: the quadcopter robot, the hexapod robot, and the ground control station (GCS). A visual representation of the architecture is presented in Figure \ref{fig:system-architecture}. The quadcopter robot consists of two primary sub-systems: the flight control system and the custom add-on. The flight control system is responsible for controlling the flight of the quadcopter and may include tele-operation, obstacle avoidance, and vision systems depending on the specific quadcopter being used. For our approach, the DJI Phantom 4 Pro drone was selected due to its cost-effectiveness, size, indoor suitability, and precise controllability. The operator at the GCS is able to tele-operate the quadcopter robot using its built-in remote controller. The custom add-on named DroneViz, which is specifically designed with a plug-and-play approach includes mapping and victim detection capabilities and is mounted on the DJI Phantom 4 Pro drone.

Primary inputs to the DroneViz come from Intel Realsense D435 camera and 9 DoF IMU. These sensors provide visual and inertial data that are used by the localisation and map creation module to perform visual-inertial and feature-based SLAM techniques to create a 3D map graph of the unknown environment. This generated map is transmitted to the GCS for the visualisation and marking of victims. In addition, the live compressed visual feed is streamed over WiFi from the Droneviz to the GCS for tele-operation and victim detection. This enables the operator to remotely control the quadcopter robot and detect victims in real-time.

The hexapod Rex is developed using the PhantomX hexapod robot kit by Trossen Robotics and has the same visual-inertial sensory system as DroneViz. The navigation module of Rex obtains the processed map with target locations from the GCS. The map is then used for 3D localisation of Rex and for autonomous navigation. The localisation of Rex is visualised in real-time at the GCS for monitoring. Additionally, the navigation module handles manual control override, navigation stack controls, and other behaviour types of Rex. Dual microphones and stereo speakers are used in a half-duplex (Walkie-talkie) manner for tele-conferencing, and a passive compartment is available for carrying a suitable medipack.

The GCS is the centralised monitoring base station which commands both the quadcopter robot and Rex. The operator can see a live visual feed with victim detection during the operation of the quadcopter robot, and can mark victim positions manually or automatically using the target identification module. After obtaining the map from the DroneViz, the map is processed and visualised as a point cloud for the operator to inspect and identify whether further exploration is needed or to return the quadcopter robot. During Rex's operation, the GCS sends the map to Rex with the target goal locations, and the GCS shows a visual indication of the live location of Rex in the generated 3D map.

 The system is built upon Robot Operating System (ROS) and the ROS Melodic version is used because of its compatibility with NVIDIA Jetson Nano and high community support. To achieve independence and robustness, a Multi-ROS communication network architecture is implemented. This allows each robot to run its own ROS master and namespace and use ROS topics to communicate between the ROS masters. This approach enables the subsystems to sync only necessary nodes and topics, becoming robust to connection dropouts in disaster scenarios and achieving scalability.

\vspace{-3pt}

\subsection{DroneViz: Custom Add-on} \label{droneaddon} 

\vspace{-10pt}

\begin{figure}[H]
  \centering
  \includegraphics[width=0.87\linewidth]{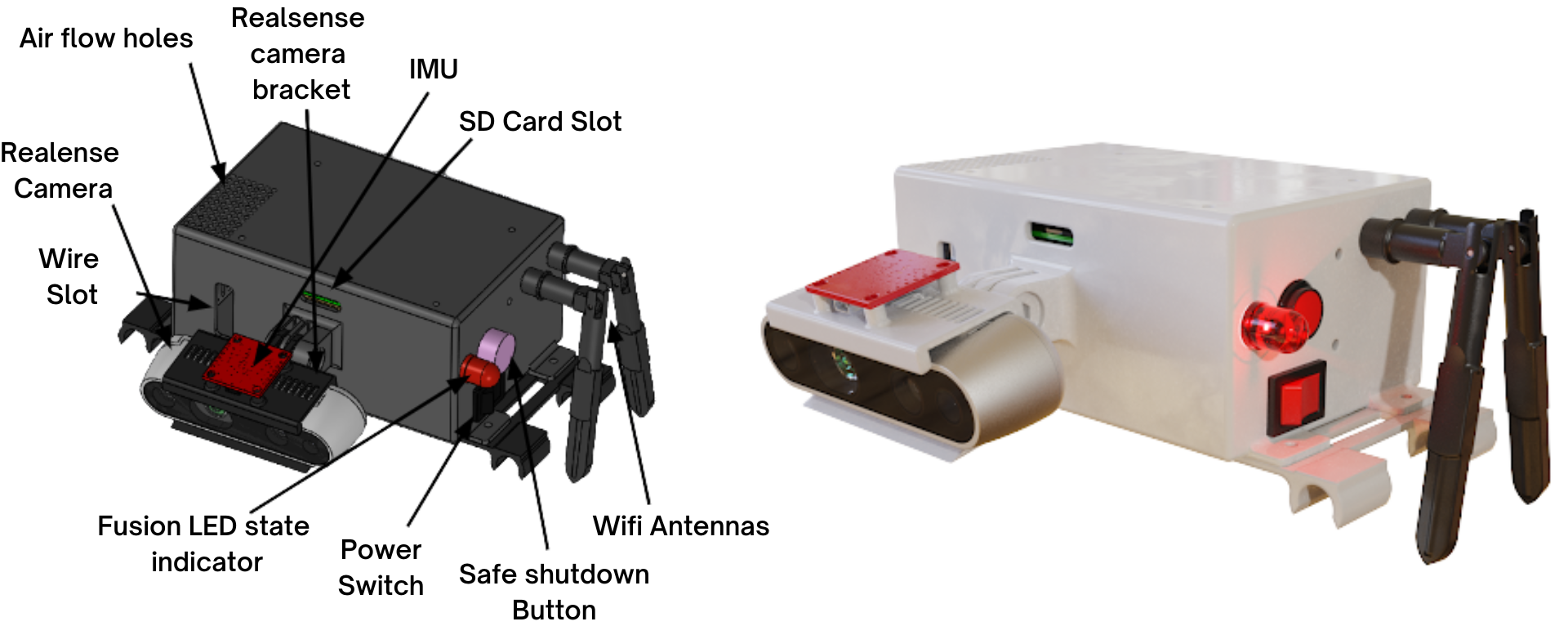} 
  \vspace{-4pt}
  \caption{DroneViz}
  \label{fig:DroneViz V2.0}
\end{figure} 

\vspace{-13pt}

The aerial robot's main duty is to map an area and locate targets while streaming live footage to the control station. DroneViz, depicted in Figure \ref{fig:DroneViz V2.0}, features safety protocols, specialized design aspects, and specific capabilities. The NVIDIA Jetson Nano 4GB board serves as the DroneViz's brain, while WiFi 802.11ac 5GHz is used for communication with the GCS. To supply power to the core device, a 3-cell lithium polymer battery with a 2200 mAh capacity is used, along with a 5V/4A step-down converter module. The core device has a safety shutdown switch and RGB LED indicators for visual cues. The Realsense D435 by Intel and ICM 20948 IMU by Sparkfun Industries serve as DroneViz's primary sensors. The DroneViz model was created utilizing the Solidworks software, where the components were meticulously positioned to ensure that the centre of gravity of the entire DroneViz was not impacted. The placement of components was tested and validated using Solidworks simulations, which ensured that the model was optimized for stability and balance when attached to an aerial robot. The final prototype was 3D printed using acrylonitrile butadiene styrene (ABS). DroneViz costs around US$\$$350 which is much less than the other methods that require building a drone with custom components.

In our selected approach, the initial search of the disaster environment is performed by an aerial robot. The aerial robot provides a unique perspective of the environment from a higher altitude, which cannot be replicated by the hexapod robot. Therefore, the use of 2D mapping would be insufficient for the hexapod to localize itself on the map provided by the aerial robot. As a result, a feature-based 3D mapping algorithm called Real-Time Appearance Based Mapping (RTAB-Map)\cite{Labbe2018} is selected as the preferred approach, allowing the hexapod to localize itself on the map by matching features, even though the viewpoints are different for the two robots.

\vspace{-13pt}

\begin{figure}[H]
    \centering
    \includegraphics[width=0.32\textwidth]{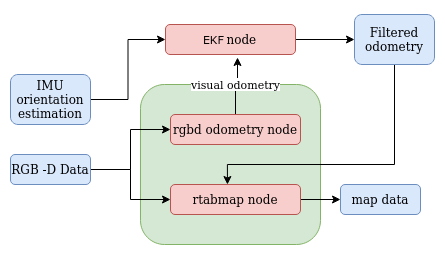} 
    \vspace{-3pt}
    \caption{Diagram for mapping with visual-inertial odometry}
    \label{fig:rtabmap_flow_vio}
\end{figure}

\vspace{-11pt}

The system depicted in Figure \ref{fig:rtabmap_flow_vio} utilizes RGB-D data acquired from an Intel Realsense D435 camera to produce visual odometry estimates. These estimates are subsequently used by the mapping node to construct a 3D map of the environment. We achieved 7Hz visual odometry estimates on the Jetson Nano, which met our needs for 3D mapping. However, in order to enhance the reliability and accuracy of our odometry estimates, and to increase the odometry rate, we investigated the use of a visual-inertial odometry approach. This approach fused visual odometry with attitude estimates from an IMU using an EKF. By combining both visual and inertial information, we achieved a more accurate and robust odometry estimate with an increased update rate of 15Hz to roll, pitch, and yaw.

To obtain an accurate attitude estimation using IMU raw data, we used a quaternion-based complementary filter as it gives more accurate estimates with a faster convergence rate when compared with EKF-based approaches. If only the gyroscope and accelerometer data are provided to the filter, it corrects only the roll and pitch of the predicted attitude estimate. If magnetometer data is also provided, yaw can be corrected as well. Hence, our complimentary filter gives attitude estimation in quaternion form from fusing gyro-scope data with accelerometer and magnetometer data in the form of two delta quaternions in an additional correction step (multiplicative approach) as shown in equation \ref{second_step}.

Here, the initial estimate computed from gyroscope data is depicted as ($q_{\omega}$), and delta quaternions derived from the accelerometer and magnetometer are depicted as $\Delta q_{acc}$, $\Delta q_{mag}$ respectively. This $\Delta q_{mag}$ is obtained by performing a rotation only about the global z-axis after aligning the global x-axis into the positive direction of the magnetic north pole. Therefore, with this formulation, such rotation does not affect the roll and pitch components even in the presence of magnetic disturbances, limiting the influence only to the yaw angle.

\vspace{-5pt}

\begin{equation}
    ^{L}_{G}q =  ^{L}_{G}q_{\omega}  \Delta q_{acc} \Delta q_{mag} 
    \label{second_step}
\end{equation}

\vspace{-2pt}

Figure \ref{fig:rtabmap_flow_vio} shows the implemented visual-inertial mapping approach for 3D mapping. Visual odometry (VO) output of RTAB-Map and attitude estimation (AE) from IMU data were fused using an EKF to obtain visual-inertial odometry. In this approach, the standard Jacobian matrix of the observation model of an EKF is replaced by a new matrix (H) to allow partial updates of the state vector. Therefore, this allows using sensors that do not measure all the variables in the state vector, which is usually the case. In our case, when measuring only VO and AE, H becomes a 2 by 12 matrix of rank 2 as shown in Table \ref{tab:state-vector}, with its only non-zero values existing in the columns of the measured variables.

\vspace{-6pt}

\begin{table}[h]
    \centering
    \caption{Matrix H}
    \label{tab:state-vector}
    \begin{tabular}{c|cccccccccccc|c}
        \cline{2-13}
        & $x$ & $y$ & $z$ & $\Phi$ & $\theta$ & $\psi$ & $x'$ & $y'$ & $z'$ & $\Phi'$ & $\theta'$ & $\psi'$ \\
        \hline
        \multicolumn{1}{|c|}{VO} & 1 & 1 & 1 & 1 & 1 & 1 & 1 & 1 & 1 & 1 & 1 & 1 \\
        \multicolumn{1}{|c|}{AE} & 0 & 0 & 0 & 1 & 1 & 1 & 0 & 0 & 0 & 0 & 0 & 0 \\
        \hline
    \end{tabular}
\end{table}

\vspace{-5pt}

Then the generated visual-inertial odometry is fed into the mapping node of the RTAB-Map to generate the 3D map of the environment in real-time. The major benefit of this approach was the ability to improve the odometry rate. The recommended odometry publishing rate for 3D mapping with RTAB-Map is around 15Hz. However, using Jetson Nano, we could only obtain visual odometry at a rate of around 7Hz. As this sensor fusion approach improved the odometry (visual-inertial odometry) rate to the recommended level, we could proceed with indoor mapping faster with fewer distortions. However, the major issue of this approach is that visual odometry works as the primary measurement source of the EKF. Out of all 12 variables of the state vector, only 3 variables (roll, pitch, yaw) are updated using the attitude estimation. Therefore, in case of visual odometry failure, the fusion method is unable to give accurate odometry data which leads to occur distortions in the map. The best way to tackle this issue is to use a tracking camera like  Intel RealSense T265 which can directly provide visual-inertial odometry. Nevertheless, we managed this issue by controlling the mapping node in instances where the visual odometry is lost.  Simply, whenever the visual odometry is lost during the mapping process, the RTAB-Map node would pause the mapping until the visual odometry gets back on track.

\subsubsection{Victim Detection}

Based on the literature survey, seven physical parameters (voice, body heat, scent, motion, skin colour, face/body shape, respiration) were identified to detect victims, and among them, body heat and face/body shape methods were selected for our experiments. Three sensors were chosen to conduct these experiments, namely PIR sensor, infrared camera, and Intel RealSense D435 camera. Out of the three sensors, Intel RealSense D435 was selected due to its wider field of view, high detection range, and high robustness. As to use with the RealSense D435 camera, the YoloV4 model, which supports ROS and provides high FPS for real-time object detection with decent accuracy, was chosen for face/body shape detection.

\vspace{1pt} 

\subsubsection{Pointing Victim Poses in Map}

 The victim detection model running at the GCS which leverages the feed from RealSense D435 data will help the controller to identify victims in a disaster area. The controller is capable of manually marking the victim poses using a specific button in the GCS dashboard or can enable automatic marking at the beginning of the mission. The victim marking methodology will use the robot's position and orientation details to mark an arrow pointing to the victim on the map which is detected through the YoloV4 model. This feature is useful for effective navigation and rescue missions.

\vspace{-4pt} 

\subsection{Rex: Hexapod Robot} \label{hexapod}

\vspace{-2pt}

We have designed a hexapod robot using Phantomx-AX chassis with 18 Dynamixel 12A motors, powered by a 5200 mAh battery. Two Jetson Nano boards have been used for locomotion, localization, navigation tasks, and tele-conferencing with a victim using a speaker and mic. It can carry payloads with a separate enclosure which has been attached to the rear of the robot and provides a live video feed for the operator to observe. Rex weighs 3.51kg and consumes 3.02A when traversing using the tripod gait. Additionally, it can operate for around 1.5 hours with the installed onboard battery.

\subsubsection{Control Aspect of Rex }

We developed a kinematic model of Rex using the DH convention and simulated it in the Gazebo physics engine in ROS by leveraging \cite{openshc_paper}. After validating, we implemented the locomotion engine on Jetson Nano for real-world use. The locomotion engine consists of a trajectory generator, tip pose generator, and a leg controller, customized and fine-tuned for Rex. We implemented and tested four gait patterns: wave, ripple, amble, and tripod. To suit targeted operation environments, we added several additional capabilities such as leg manipulation, and manual body posing on top of the locomotion engine. The Logitech F710 wireless gamepad enables manual control when the Rex reaches the victim location, allowing further inspection of the area. Furthermore, it enables the user to adjust the tip positions of selected legs, which is helpful for clearing objects in the robot's path. Rex's Intel RealSense D435 camera is currently unable to rotate or tilt, which limits its field of view. To address this and enable the robot to traverse in confined spaces, we added manual body posing capability. The robot can control its body along x-y axes, pitch-roll axes, and z axis.

\subsubsection{Autonomous Navigation}

We implemented a customized navigation stack in ROS that aligns with Rex. The quadcopter's map is fed to the navigation stack, which uses A* algorithm as the global planner and DWA planner as the local planner. Rex can handle dynamic obstacles by updating costmaps with data from the Intel RealSense D435 camera. Localization of Rex in the map provided by DroneViz is performed using feature mapping.

\subsubsection{Tele-conferencing and Payload Delivery}

After the evaluation of several methodologies, we have used a Waveshare audio card that can be conveniently installed on the Jetson Nano board. The system was implemented on Rex using the half-duplex walkie-talkie method to reduce communication complexity.
An enclosure has been created with a sliding door open mechanism that can be used for payload delivery tasks. The design is a prototype, and the volume can be changed by users to fit their needs while staying under the maximum weight constraint of 200g when using the amble gait.

\vspace{-5pt}

\subsection{Ground - Aerial Collaboration} \label{collab}

\vspace{-3pt}

Rex can autonomously navigate to target locations by using the 3D map with target locations generated by DroneViz. Before navigation, Rex localizes itself in the map using the localization mode in the RTAB-Map node. Even though the viewpoints of Rex's and DroneViz's cameras differ due to the height difference, RTAB-Map can handle them by using feature mapping.

\vspace{-5pt}

\section{Results}
\vspace{-2pt}

In this section, we present the evaluations of the proposed system considering DroneViz and Rex.

\vspace{-6pt}

\begin{figure}[H]
  \centering
  \includegraphics[width=0.7\linewidth]{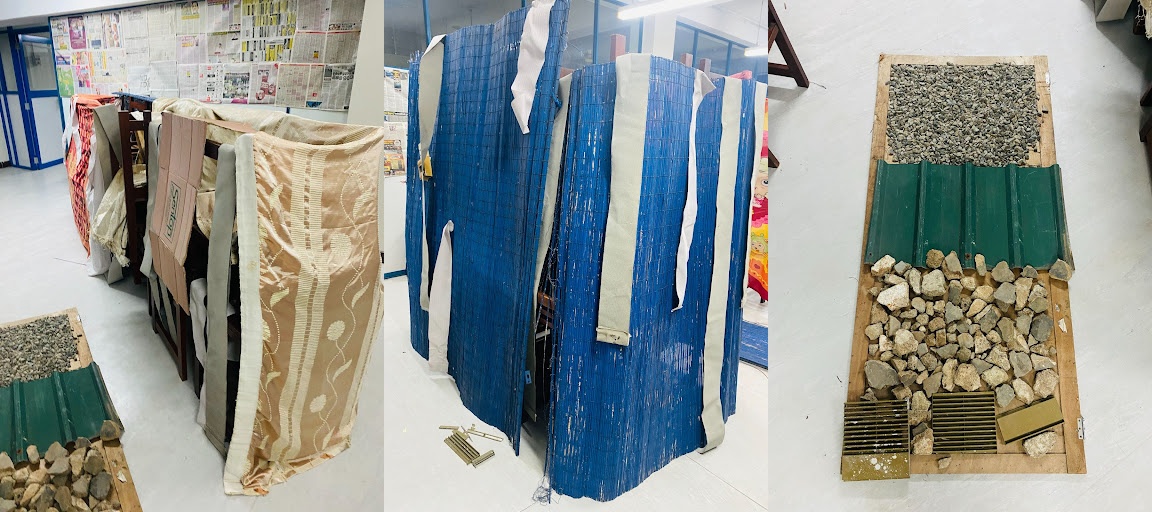}  
  \caption{Images taken from the created disaster environment}
  \label{fig:Disaster Environment}
\end{figure}

\vspace{-10pt}

These evaluations were conducted in an indoor environment (Figure \ref{fig:Disaster Environment}) that includes walls of different heights, different types of objects, and terrains. This environment was designed to simulate a real-world disaster scenario where the system could be deployed. Additionally, we have performed simulations for Rex in the Gazebo simulator to ensure the algorithms are working as expected before testing in the real world.

\subsection{DroneViz Evaluations}

\vspace{-2pt}

\subsubsection{Tests and Parameters}

The DroneViz's hardware design was validated through various tests, including a maximum load test and a weight test. The maximum load was found to be 2A and the total weight of the DroneViz was 650g. This has a longer operation time of 1 hour compared to the selected drone's (DJI Phantom 4 Pro) flight time of around 30 minutes.

\subsubsection{Accuracy of Mapping}

\begin{figure}[H]
  \centering
  \includegraphics[width=0.95\linewidth]{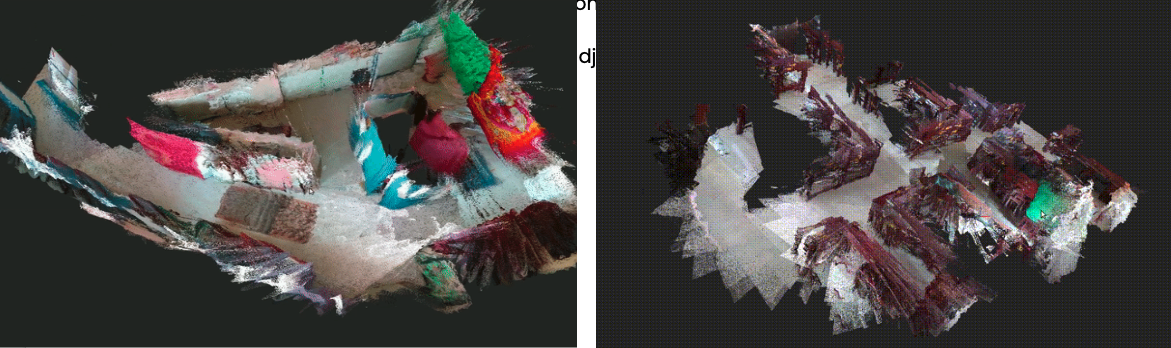}  
  \caption{Improved 3D maps}
  \label{fig:improvement_mapping}
\end{figure}

\begin{figure}[H]
  \centering
  \includegraphics[width=0.73\linewidth]{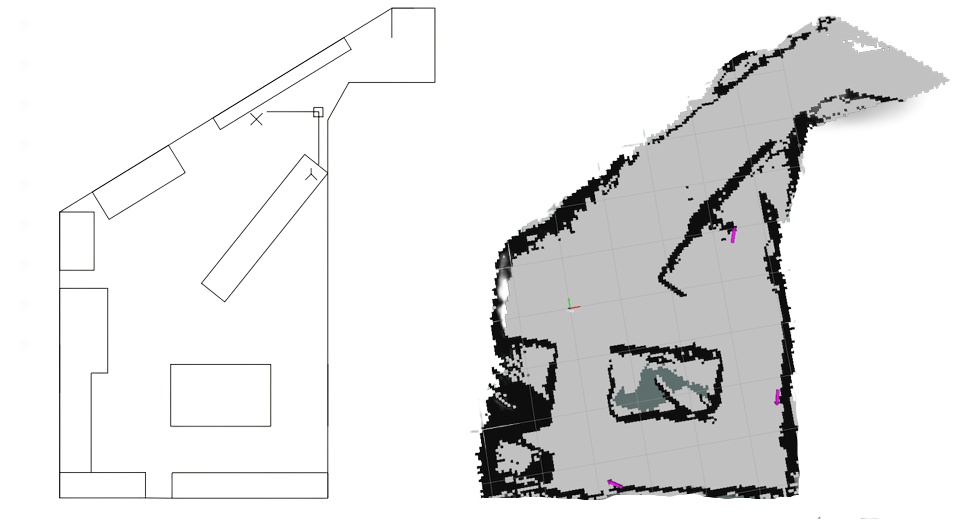}  
  \vspace{-5pt}
  \caption{Comparison of ground truth with 2D test map}
  \label{fig:compare-maps}
\end{figure}

Figure \ref{fig:improvement_mapping} shows two 3D maps generated using visual-inertial odometry. Using this approach improved the mapping accuracy when compared with the maps generated using only visual odometry, and the accuracy was further increased with improved visual-inertial odometry by activating internal digital low pass filters of the ICM-20948 IMU which made the output raw data free of high-frequency noise. 
A ground truth 2D map was created and compared with the 2D map generated by DroneViz as shown in Figure \ref{fig:compare-maps}.

\vspace{-7pt}

\vspace{-13pt}

\vspace{-3pt}

\vspace{-10pt}

\subsubsection{Victim Detection Results}

\begin{figure}[H]
  \centering
  \includegraphics[width=1\linewidth]{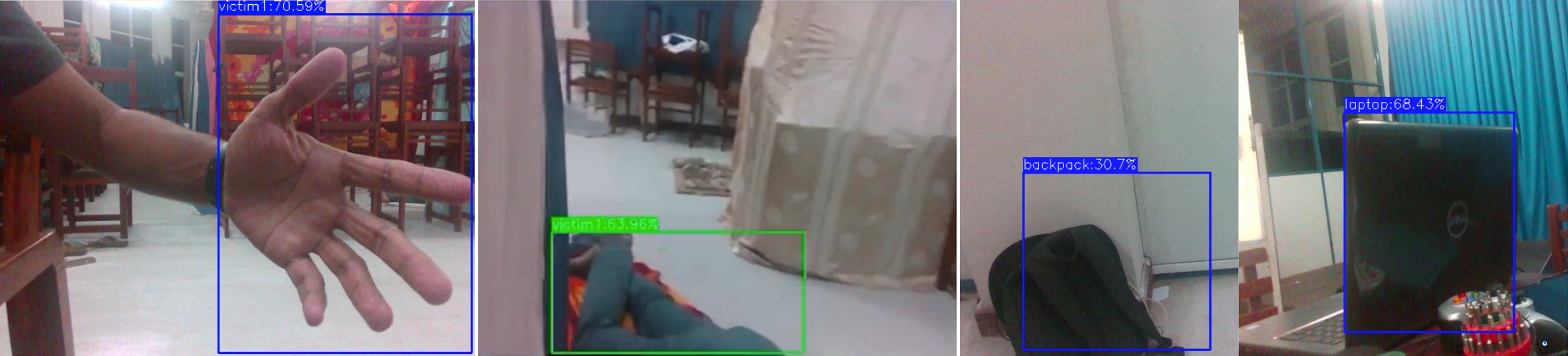} 
  \vspace{-14pt}
  \caption{Detecting victims in extreme cases and related objects}
  \label{fig:detect-victims}
\end{figure}

The performance of the YOLOv4 model was evaluated on the ground control station, using an ASUS TUF F5 laptop. The robustness of the YOLOv4 model is visualized in the figures, and the C++ based Darknet Yolo implementation was used for inference. As shown in Figure \ref{fig:detect-victims}, the model was able to identify victims even if only a part of the body was within view, and it can identify other objects which make suspicious of whether a victim is nearby.

\vspace{-9pt}

\vspace{-10pt}

\begin{figure}[H]
  \centering
  \includegraphics[width=0.9\linewidth]{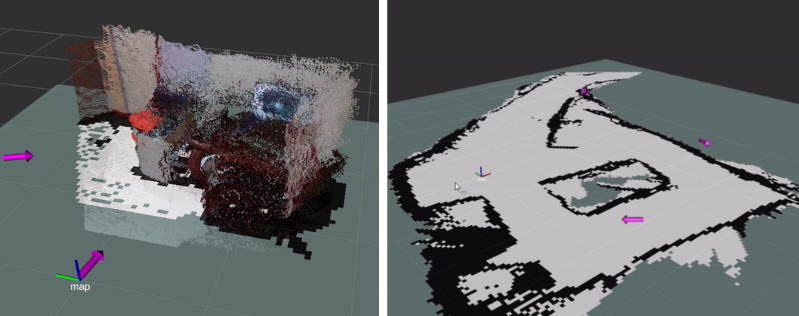} 
  \caption{Marking victims' poses}
  \label{fig:vic_poses}
\end{figure}

Marking victims' poses on the map plays an important role in the system. Figure \ref{fig:vic_poses} shows some sample maps, marked with victim poses. Arrowheads show the directions of victims from the point where the victim is detected.

\vspace{-8pt}

\vspace{-16pt}

\subsection{Rex Evaluations}

\vspace{-3pt}

Given the intended use of Rex in challenging environments, it is critical to validate its capabilities. To achieve this, a series of tests were conducted and can be broadly classified as current tests, torque tests, and traversability tests. 

\hspace{2pt}
\vspace{-10pt}

\subsubsection{Current Tests}

In order to determine the maximum operation time and select a suitable battery, the current consumption of Rex was measured and plotted using DC power supplies. The goal was also to identify body heights that minimize current consumption during stance and gait stages. After conducting several current tests, the lowest current consumption was observed at a body height of 14cm during the stance and 12cm during the gait stage. These parameters were programmed into Rex to minimize current consumption.

\hspace{2pt}
\vspace{-10pt}

\subsubsection{Torque Tests}

\begin{figure}[H]
    \centering

    \begin{subfigure}[b]{0.44\columnwidth}
        \centering
        \includegraphics[width=0.95\linewidth]{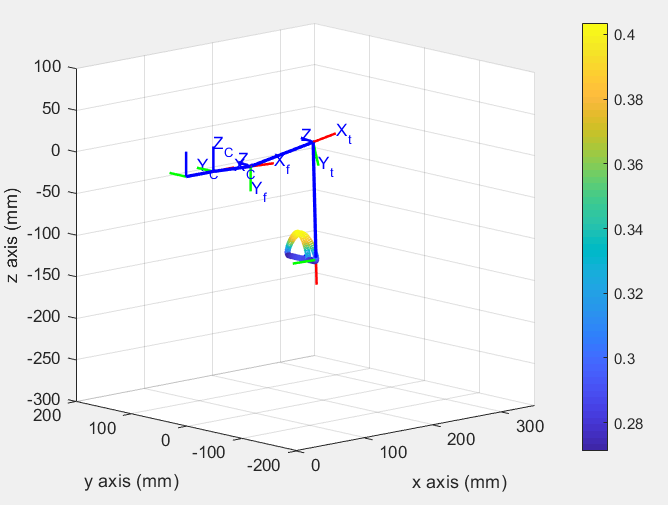} 
        \caption{During tripod gait
}
        \label{fig:B}
    \end{subfigure}  
    \begin{subfigure}[b]{0.42\columnwidth}
        \centering
        \includegraphics[width=0.95\linewidth]{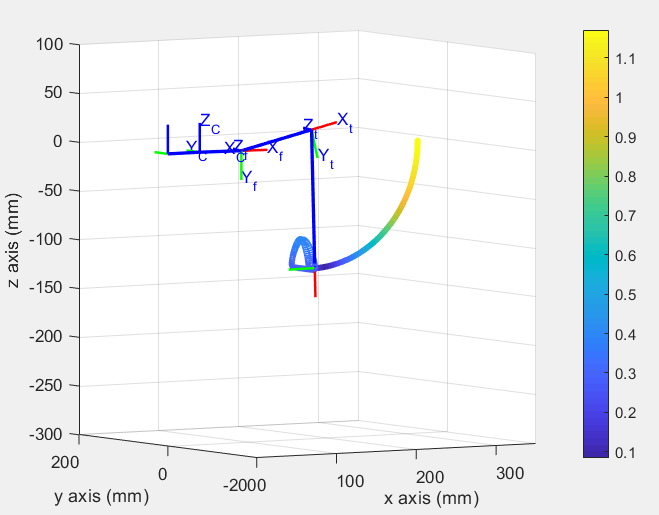}
        \caption{During stance to walking}
        \label{fig:B}
    \end{subfigure}
    \vspace{-3pt}
    \caption{Torque test results}
    \label{fig:currenttest}
\end{figure}

To ensure the safety and maximum performance of legged robots, it is necessary to determine their maximum payload capacities. The Dynamixel 12A motors used in Rex have a stall torque of 1.5Nm, and calculations were performed for the tripod gait, which is the worst-case scenario when it comes to payload bearing. Analysis as depicted in Figure \ref{fig:currenttest} showed that Rex can safely carry all its components, even in the tripod gait, with a safety margin of 25-33 $\%$ from the stall torque. Changing the gait to amble can increase the payload capacity by 180-200g, and further increases are possible with slower gaits like wave or ripple.

\vspace{-8pt}

\vspace{-10pt}
\subsubsection{Traversability Tests}

\begin{figure}[H]
  \centering
  \includegraphics[width=0.75\linewidth]{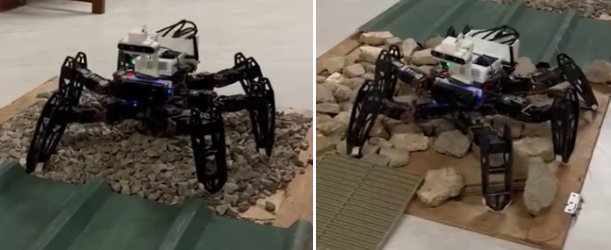} 
  \caption{Rex traversing in different terrains}
  \label{fig:travertests}
\end{figure}

To validate the performance in different and difficult terrains, Rex has been rigorously tested in various terrains as shown in Figure \ref{fig:travertests}.

\vspace{-7pt}

\vspace{-16pt}

\section{Conclusion}

\vspace{-3pt}

The proposed collaborative system involves an aerial and ground robot to explore and map environments, identify targets, deliver a medi-pack, and tele-conference with a victim. The system is low-cost, effective, and scalable to a team of robots, making it useful in search and rescue missions. The solution is easily adaptable to any off-the-shelf drone capable of carrying the drone add-on. The use of a six-legged robot provides benefits in stability and leg manipulation, and the camera mounted on it is useful for exploration tasks.

\vspace{-2pt}

\section{Limitations and Future Work}

\vspace{-2pt}

The research aimed to improve odometry in low-light conditions by combining attitude estimation from an external IMU with visual odometry. However, this approach did not solve the problem of losing the primary odometry measurement source. To address this limitation, we plan to add another primary odometry measurement unit for improved mapping in darker and featureless environments. We also plan to upgrade the hexapod's motors and evaluate their methodology in larger outdoor environments. Moreover, the proposed solution will be expanded to research areas such as multi-robot mapping and exploration. Finally, we aim to develop DroneViz as a commercial low-cost and low-weight drone add-on.

\vspace{-3pt}

\section{Acknowledgements}

\vspace{-2pt}
We acknowledge the resources provided by the Commonwealth Scientific and Industrial Research Organisation (CSIRO), Australia, during this project.

\vspace{-1pt}

\bibliographystyle{IEEEtran} 
{\footnotesize
\bibliography{refs} 
}

\end{document}